\documentclass[10pt, a4paper]{article}
\usepackage{lrec2022} 
\usepackage{multibib}
\newcites{languageresource}{Language Resources}
\usepackage{graphicx}
\usepackage{tabularx}
\usepackage{soul}
\usepackage{multirow}
\usepackage{titlesec}
\titleformat{\section}{\normalfont\large\bfseries\center}{\thesection.}{1em}{}
\titleformat{\subsection}{\normalfont\SmallTitleFont\bfseries\raggedright}{\thesubsection.}{1em}{}
\titleformat{\subsubsection}{\normalfont\normalsize\bfseries\raggedright}{\thesubsubsection.}{1em}{}
\renewcommand\thesection{\arabic{section}}
\renewcommand\thesubsection{\thesection.\arabic{subsection}}
\renewcommand\thesubsubsection{\thesubsection.\arabic{subsubsection}}

\usepackage{epstopdf}
\usepackage[utf8]{inputenc}

\usepackage{hyperref}
\usepackage{xstring}

\usepackage{color}
\usepackage{tipa}

\usepackage{booktabs}

\title{Preparing an Endangered Language for the Digital Age: The Case of Judeo-Spanish}

\name{Alp Öktem$^1$, Rodolfo Zevallos$^{1,2}$, Yasmin Moslem$^3$, Güneş Öztürk$^1$, Karen Şarhon$^4$} 
\address{$^1$Col$\cdot$lectivaT, Barcelona, Spain
  $^2$Universitat Pompeu Fabra, Barcelona, Spain\\
  $^3$Dublin City University, Dublin, Ireland
  $^4$Sephardic Center of Istanbul, Istanbul, Turkey\\
         alp@collectivat.cat, rodolfojoel.zevallos@upf.edu, yasmin.moslem@adaptcentre.ie\\ ozgurgunes@collectivat.cat, karensarhon@gmail.com\\}

\abstract{
We develop machine translation and speech synthesis systems to complement the efforts of revitalizing Judeo-Spanish, the exiled language of Sephardic Jews, which survived for centuries, but now faces the threat of extinction in the digital age. Building on resources created by the Sephardic community of Turkey and elsewhere, we create corpora and tools that would help preserve this language for future generations. For machine translation, we first develop a Spanish to Judeo-Spanish rule-based machine translation system, in order to generate large volumes of synthetic parallel data in the relevant language pairs: Turkish, English and Spanish. Then, we train baseline neural machine translation engines using this synthetic data and authentic parallel data created from translations by the Sephardic community. For text-to-speech synthesis, we present a 3.5 hour single speaker speech corpus for building a neural speech synthesis engine. Resources, model weights and online inference engines are shared publicly. 
 \\ \newline }

\begin{document}

\maketitleabstract

\section{Introduction}

In this paper, we present our ongoing language technology-related efforts for preparing Judeo-Spanish to the digital age. We embark upon creating open language corpora and tools that would serve for language documentation, assisting language learners and development of advanced applications. We focus on two main tools, machine translation (MT) and text-to-speech synthesis (TTS). In our extremely low-resource setup, we get use of Judeo-Spanish’s proximity to Spanish by using transfer learning methodologies. For MT, we build a rule-based machine translation engine that allows us to convert Spanish text to Judeo-Spanish. Using this system, we create large synthetic pre-training data from publicly available English, Turkish and Spanish parallel corpora and train neural machine translation systems. For TTS, we do transfer learning from pre-trained Spanish and English engines using a small single-speaker speech corpus. During the development of these tools, we have packaged various types of raw resources into training-ready language data and models and shared them in our project's data portal \textit{Ladino Data Hub}\footnote{\url{http://data.sefarad.com.tr}}. The complete list of output of this work can be presented as follows: 
\begin{enumerate}
\item A monolingual news corpus, 
\item Authentic and synthetic parallel corpora in English, Spanish and Turkish paired with Judeo-Spanish, 
\item A Spanish to Judeo-Spanish rule-based machine translation system,
\item Neural machine translation models between Judeo-Spanish and English, Spanish and Turkish,
\item A 3.5 hour single speaker speech corpus,
\item Neural network-based speech synthesis model,
\item Web application for MT and TTS\footnote{\url{http://translate.sefarad.com.tr}}.
\end{enumerate}

\section{Background}

Judeo-Spanish, also referred to as Ladino or Judezmo (ISO 639-3 $lad$), is a descendant of old Castilian Spanish from the 15th century \cite{sefardiweb}. It is the historical and predominant language of the Sephardic Jews, who were expelled from their homes by the Spanish Inquisition (1492) and welcomed into the Ottoman Empire, where they retained the language, as well as France, Italy, the Netherlands, Morocco and England, where they shifted to the dominant language. It has traces of numerous Iberian languages of the 15th century like Old Aragonese, Astur-Leonese, Old Catalan, Galician-Portuguese and Mozarabic with Castilian Spanish forming its basis vocabulary \cite{minervini}. 
After 530 years, Judeo-Spanish still survives as a language of Ottoman Sephardic Jews in more than 30 countries, with most speakers residing in Israel. Although it has survived and evolved over the centuries, it is currently classified as a severely endangered language by UNESCO \cite{unesco}. 

The digital age has a direct effect on endangered languages like Judeo-Spanish. There is currently a growing digital divide between languages with sufficient resources and languages with fewer resources, further exacerbating the danger of digital extinction for them \cite{kornai}. For the dominant languages the process of generating artificial intelligence tools is much easier due to their large web-presence. However, many marginalized languages do not have sufficient material and human resources to power the creation of such tools. Lack of state support, public visibility, as well as societal and institutional oppression are direct causes of these languages being deprioritized in the digital spaces of today \cite{nekoto2020participatory}. 

The Sephardic community of Turkey has been active in promoting their language heritage in many ways. These include: publishing the only newspaper in the world entirely in Judeo-Spanish \textit{El Amaneser}, giving language lessons, writing and performing plays in Judeo-Spanish, creating language learning content, collecting speech corpora and publishing dictionaries, music albums and books. 

The aim of this work is to build data-centric technology for Judeo-Spanish for it to gain digital ground. Besides building new and compiling already existing corpora for this purpose, we create first machine translation and text-to-speech synthesis systems for the language. Machine translation makes it possible for the language to be interpretable by non-speakers and is also proposed as a way of language documentation \cite{bird-chiang-2012-machine}. It is now also considered as an attractive tool for many language learners in addition to dictionaries and thesauri \cite{duke}. Even though it is difficult to obtain high performance in low resource settings, it has been used to strike interest in language and collect translations and corrections from the community. The second language tool we focus on, text-to-speech synthesis (TTS), makes it possible building of tools like virtual assistants and screen readers. In the context of language learning, one can learn how a certain word or sentence is pronounced in a language without the help of an instructor or a speaker. 

\section{Judeo-Spanish Resources}

We explain our various data compilation efforts in this section. All data presented are published with \textit{CC BY-SA 4.0 license}\footnote{\url{http://creativecommons.org/licenses/by-sa/4.0/}} on Ladino Data Hub. We also provide the scripts we have used in developing these resources with GPL-licenses for facilitating expansion and reproducibility\footnote{\url{http://github.com/CollectivaT-dev/judeo-espanyol-resources}}.

\subsection{Monolingual text corpus}

Text corpora have been used both in language technology and in linguistic research. They are an essential part of creating statistical language models that are used in applications such as optical character recognition, handwriting recognition, machine translation and spelling correction.

For this task, we automatically scraped the articles published in the weekly online newspaper \textit{Şalom}\footnote{\url{http://www.salom.com.tr}}. As of now, we have collected $397$ articles totaling to $176,843$ words.

\subsection{Parallel corpus}
\label{sec:authentic}

The type of data that is needed to build a MT system is parallel data, which consists of a collection of sentences in a language together with their translations. We have only detected two publicly available corpora of Judeo-Spanish in the commonly used OPUS portal\footnote{\url{http://opus.nlpl.eu/}}: Wikimedia corpus consisting of 18 sentences and Tatoeba corpus of 872 sentences. 

In order to expand on this set, we gathered translations made by the Sephardic Center of Istanbul. These covered topics like news articles, online shop strings, recipes and cultural event announcements. We automatically segmented the text into sentences getting use of punctuation and then manually verified alignments. We also digitized the language learning material \textit{Fraza del dia}\footnote{\url{http://sefarad.com.tr/judeo-espanyolladino/frazadeldia/}}, where daily a Judeo-Spanish phrase is presented with their translations in another language. The sizes of parallel corpora created for each language pair is listed in Table \ref{tab:parallel}.

\begin{table}[htbp]
\begin{tabular}{lcc}
\cline{1-3}
\begin{tabular}[c]{@{}l@{}}\textbf{Language pair}\\ (Judeo-Spanish and)\end{tabular}  & \textbf{\#Sentences} & \textbf{Total \#tokens}   \\ \cline{1-3}
English                & 3333                 & 41,508  \\
Spanish                & 977                  & 12,712   \\
Turkish                & 845                  & 15,781   \\ \cline{1-3}
\end{tabular}
\caption{Parallel data compiled from Tatoeba and translations by Sephardic community.}
\label{tab:parallel}
\end{table}

\subsection{Spanish Judeo-Spanish Dictionary}
\label{sec:dict}

We developed a digital Spanish--Judeo-Spanish dictionary from the sources listed in Table \ref{tab:dictionary}. To process the dictionaries shown in Table \ref{tab:dictionary} which were in PDF format, we used the Python programming language, where we aligned the Spanish word with Judeo-Spanish word and eliminated irrelevant information like example sentences. Once the dictionaries were processed, the data were stored in a plain text file under the following structure: $\langle word$-$spanish, word$-$judeospanish \rangle$.


\begin{table}[htbp]
\begin{center}
\begin{tabular}{lc}
\hline
Dictionary   &  \# Entries \\ \hline
\begin{tabular}[c]{l@{}}Diksionaryo de Ladino \\ a Espanyol \\ \cite{dic12003} \end{tabular}        & 2523       \\ \hline
\begin{tabular}[c]{l@{}}Diksionario de \\ Djudeo-Espanyol a \\ Castellano \\ \cite{dic12009}\end{tabular}         & 4215   \\ \hline   
\end{tabular} 
\caption{Dictionaries used for the construction of the digital dictionary.}
\label{tab:dictionary}
\end{center}
\end{table}

\subsection{Single-speaker speech corpus}
\label{sec:ttscorpus}

We built a single-speaker speech corpus of 3 hours and 24 minutes to be used in the creation of Judeo-Spanish TTS system. We had our native Judeo-Spanish speaking author read 30 articles from the weekly newspaper \textit{El Amaneser}. The articles are about different topics, ranging from historical issues, current affairs, cultural events and politics. The recordings had an average length of 6 minutes. To obtain TTS training data material, we had to divide the audios into smaller segments. For this task, we developed an automatic aligner\footnote{\url{https://github.com/CollectivaT-dev/Judeo-Spanish_STT}} based on Coqui Speech-to-text \cite{coqui}. The pre-trained Spanish model performed well enough to optimize the process. Nevertheless, to ensure completely matching audio and transcription pairs, we manually verified each pair and performed corrections where needed. The resulting corpus consists of 1987 16-bit, single-channel WAV audio files sampled at 16kHz with their transcriptions.

\section{Machine Translation}

In this section, we present our experiments for Judeo-Spanish machine translation. To account for the lack of data, we first build a rule-based Spanish to Judeo-Spanish translator and then use that to obtain the data needed to train neural baseline models.

\subsection{Rule-Based machine translation}
\label{sec:rule}
In the following, we describe the procedure of our rule-based machine translation system from Spanish to Judeo-Spanish based on the dictionaries available in Table \ref{tab:dictionary}. The Python-based scripts and documentation are provided with GNU General Public License in our Github repository\footnote{\url{https://github.com/CollectivaT-dev/Espanyol-Ladino-Translation}}.

The first step in the translation process is to tokenize the input Spanish phrase, for which we use the Python library Stanza\footnote{Stanza is a collection of tools for the linguistic analysis (Tokenization, Part-of-Speech, Lemmatization, etc.) of many human languages, including Spanish. \url{https://stanfordnlp.github.io/stanza/}} \cite{qi2020stanza}. This library, in addition to tokenizing the phrase, obtains the part-of-speech (POS) and lemmas of each token. As a second step, each token is looked up in the Spanish--Judeo-Spanish dictionary. If the token is found in the dictionary, its corresponding Judeo-Spanish token is obtained, otherwise, the dictionary is searched for its lemmatized form of the token. If the lemmatized token is found in the dictionary, the corresponding Judeo-Spanish token is obtained and is conjugated according to its POS. Our method transforms a Spanish token to a conjugated Judeo-Spanish form using an algorithm based on conjugation rules specified in \cite{dic1997}. 
We also convert the verb form Present Perfect (e.g. spa.``he cocinado'') to past indefinite (e.g. spa.``cociné'' lad. ``gizi'') as the former form is not common in Judeo-Spanish. In case the lemmatized token is not found in the dictionary, it is processed by a Judeo-Spanish correction method, which follows established orthographic rules of the language\footnote{Orthographic structure of Judeo-Spanish compared to Spanish available in \url{https://github.com/CollectivaT-dev/judeo-espanyol-resources/blob/main/resources/Gramatica_Ladino.doc}}. Finally, in step three, phrase forms that do not exist in Judeo-Spanish are corrected into their right form using a phrase correction dictionary. For example, \textit{"tengo ke"} (from \textit{spa.``tengo que''} \textit{eng.``I have to''}) is corrected to \textit{``debo de''}, or \textit{"ay ke"} (from \textit{spa.``hay que''} \textit{eng.``one must''}) is corrected to \textit{``Kale''}. Some example translations are listed in Table \ref{tab:rulebased}. Automatic evaluation results are presented in Table \ref{tab:bleu}. 

\begin{table*}[]
\begin{center}
\begin{tabular}{ll}
\hline
\textbf{Spanish input} & \textbf{Judeo-Spanish translation} \\ \hline
Me gusta leer. & Me plaze meldar. \\ 
¿No has leido el libro? & No meldates el livro? \\
Bebo café turco después del almuerzo. & Bevo kafe turko despues del komida de midi. \\ 
Tengo dos niños; una hija y un hijo. & Tengo dos kriaturas; una ija i un ijo. \\ 
Tengo que cocinar para mañana. & Devo de gizar para amanyana. \\  \hline
\end{tabular}
\caption{Example translations obtained with the rule-based machine translation system.}
\label{tab:rulebased}
\end{center}
\end{table*}

\subsection{Data augmentation} 

\begin{table}[htbp]
\centering
\begin{tabular}{lr}
\textbf{ENG-SPA}       & \textbf{\#sentences }       \\ \hline
Books                  & 93,470             \\
Europarl               & 615,626         \\
News-commentary        & 49,089             \\
OpenSubtitles          & 4,652,910          \\
SciELO                 & 164,500            \\
TED2013                & 157,895            \\
WMT-News               & 14,522             \\ \hline
\textbf{TOTAL ENG-SPA} & \textbf{5,748,012}  \\ \hline \hline
\textbf{SPA-TUR}       &                   \\ \hline
EUBookshop             & 19,914             \\
GlobalVoices           & 7,461              \\
OpenSubtitles          & 4,000,000           \\
TED2020                & 370,465            \\
Tatoeba                & 28,829             \\
WikiMatrix             & 147,352            \\ \hline
\textbf{TOTAL SPA-TUR} & \textbf{4,574,021}  \\ \hline
\textbf{TOTAL SPA} & \textbf{10,322,033} \\ \hline
\end{tabular}
\caption{Publicly available parallel data used for synthetic data creation.}
\label{tab:augmentation}
\end{table}

We introduce a data augmentation method based on creating synthetic parallel data using the rule-based MT system presented in Section \ref{sec:rule}. We first collect publicly available parallel data in pairs English-Spanish and Turkish-Spanish from the OPUS collection. Then, we translate the Spanish portions into Judeo-Spanish using the rule-based MT. This yields  Turkish--Judeo-Spanish and English--Judeo-Spanish synthetic parallel data. Finally, the Spanish portions of two sets are then merged to create Spanish--Judeo-Spanish synthetic parallel data. The statistics and sources for synthetic data augmentation are listed in Table \ref{tab:augmentation}.

\subsection{Neural machine translation}

We used the OpenNMT-py toolkit \cite{klein-etal-2018-opennmt} to train the models. The model consists of an eight-head Transformer ``big'' \cite{DBLP:conf/nips/VaswaniSPUJGKP17} with six-layer hidden units of 512 unit size. It uses Relative Position Representations \cite{shaw-etal-2018-self} with a clipping distance k=16. A token-batch size of 1,024 was selected. Adam optimizer \cite{DBLP:journals/corr/KingmaB14} was selected with 4,000 warm-up steps. Trainings were performed until no further improvement was recorded in development set perplexity in the last five validations.

We used the synthetic parallel data we created as training data. As for development and test sets, we used the authentic data mixes presented in Section \ref{sec:authentic}. As English portion was about three times larger than Spanish and Turkish, we used a commercial machine translation engine to translate the extra data available for English to Spanish and Turkish and added them to the mixes. Finally, we reserved 500 sentences from Spanish and Turkish and 750 sentences from English mix as test data and used the rest as validation data during training. Development, test sets, training configuration files, subword models and training logs are provided for reproducibility\footnote{\url{https://github.com/CollectivaT-dev/judeo-espanyol-resources/tree/main/MT_devtest_configs}}. Model weights are made available in Ladino Data Hub. 

\begin{table}[htbp]
\centering
\begin{tabular}{lccc} \hline
                                    & \textbf{ENG} & \textbf{SPA} & \textbf{TUR} \\ \hline
\textbf{LAD $\rightarrow$ \textit{lang}}  & 34.96   & 47.13   & 20.14   \\
\textbf{\textit{lang} $\rightarrow$ LAD}  & 26.03   & 44.85   & 21.03   \\
\textbf{Rule-based SPA $\rightarrow$ LAD} &  -      & 45.80   &  -  \\ \hline    
\end{tabular}
\caption{Automatic evaluation results in 6 language directions and also on rule-based system. BLEU scores were calculated on lowercased output and reference with SACREBLEU toolkit with Moses tokenizer  \protect\cite{post-2018-call}. }
\label{tab:bleu}
\end{table}

We report our test set BLEU-scores \cite{papineni-etal-2002-bleu} for each translation direction in Table \ref{tab:bleu}. As future work, we will also perform human evaluations on additional data to have a fairer judgment of the translation qualities. 




\section{Text-to-Speech}

In this section, we present our experiments for the development of a Judeo-Spanish speech synthesizer. We use Glow-TTS model \cite{kim2020glow} for our experiments and the Griffin-Lim algorithm \cite{griffin1984signal} to avoid using vocoder, which we intend to develop for future research. We trained three Glow-TTS models with our 3.5 hour dataset. The first model was trained from scratch; the second and third, by fine-tuning English and Spanish TTS models. For all experiments, we do not enable the use of phonemes or the phonemizer as in \newcite{yurii2021}. We follow the settings for the mel-spectrogram of \newcite{prenger2019waveglow}.

\textbf{Training Judeo-Spanish from scratch} First, we evaluate the performance of the model trained only using our single-speaker dataset. During training, like \newcite{kim2020glow} we set the standard deviation to 1. Our model was trained for 5,000 iterations with a batch size of 32, using the Adam optimizer \cite{kingma2014adam} with the Noam learning rate program \cite{vaswani2017attention}. This required only 4 days with an 8GB NVIDIA GPU.

\textbf{Fine-Tuning from English and Spanish} To train the Glow-TTS model from the pre-trained English (ljspeech/glow-tts) and Spanish (mai/tacotron2-DDC) models \cite{coqui}, we used the same setup as the Glow-TTS model trained only with Judeo-Spanish data, but added the use of the phonemes and phonemizer corresponding to each language from the pre-trained models. Each model required only 4 days with an 8GB NVIDIA GPU.

\begin{figure*}[ht]
\begin{center}
\label{fig:translator}
\includegraphics[width=1.0\textwidth]{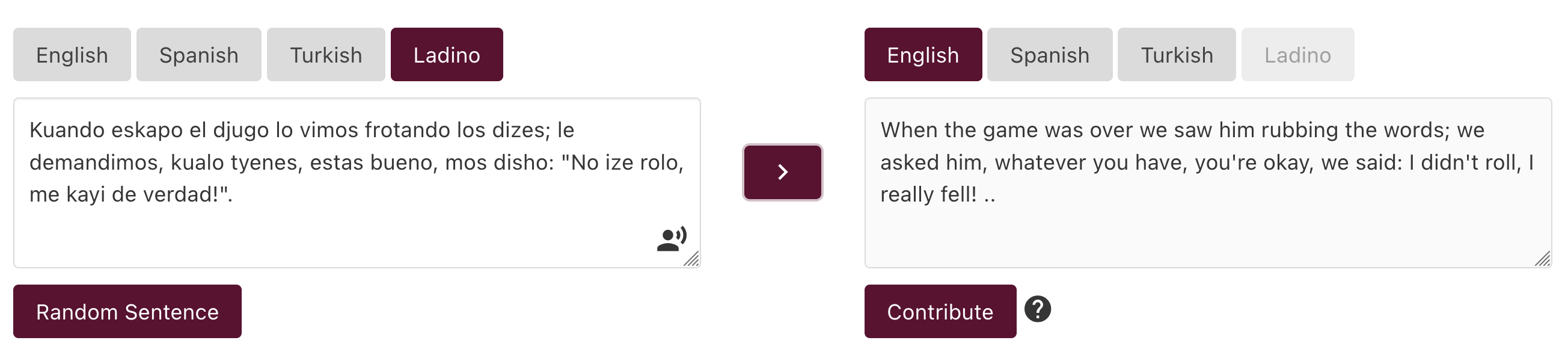}
\caption{Web application for MT and TTS available in \url{http://translate.sefarad.com.tr} }
\end{center}
\end{figure*}

\subsection{Evaluation}

Our best model was the one where fine-tuning was applied to the pre-trained English model, achieving a much better intelligibility and naturalness than the other models, reducing even more the ``metallic'' sound that appears in some consonants. On the other hand, the Spanish fine-tuned model did achieve a better naturalness for Judeo-Spanish phonemes but did not achieve a good intelligibility, perhaps due to the amount of training data. Likewise, the from-scratch model did achieve an excellent naturalness in the phonemes but a very poor intelligibility. 

We selected the best performing model (fine-tuned from English) for human evaluation. We used Mean opinion score (MOS) of intelligibility and naturalness with a 5-point scale: 5 for excellent, 4 for good, 3 for fair, 2 for poor and 1 for bad. An evaluation survey consisting of ten out-of-corpus samples were published in a closed Ladino speaker community of Istanbul and 12 native speakers participated. The average scores among ten samples are listed in Table \ref{tab:tts}. 
The configurations of all pre-trained models as well as audio samples are made available online for reproducibility \footnote{\url{https://github.com/CollectivaT-dev/Ladino_TTS}}. Model weights are shared in Ladino Data Hub.

\begin{table}[htbp]
\begin{center}
\begin{tabular}{lcc}
\hline
Model & Intelligibility & Naturalness\\ \hline
Glow-TTS \\ (f.t. on English)&4.04& 3.61 \\ \hline
\end{tabular}
\caption{Judeo-Spanish text-to-speech system evaluation results for intelligibility and naturalness (MOS)}
\label{tab:tts}
\end{center}
\end{table}

\section{Web Application}

Our web application for serving the machine translation and speech synthesis systems can be seen in Figure \ref{fig:translator}. It allows translation between English, Spanish, Turkish and Ladino and makes it possible to listen to sythesized Ladino text. For Spanish, we integrated the rule-based system translating to Ladino and our model translating to Spanish. For the rest of the translation directions, we chained open source OPUS-MT translation models \cite{TiedemannThottingal:EAMT2020} to these two systems to get translation to and from English and Turkish.

We also added a participation feature to make Judeo-Spanish speakers be part of future developments. By clicking the "Contribute" button, users can correct the translations and then submit to our database to be stored as parallel data for future trainings. 

\section{Conclusion}

In this work, we introduced baseline systems of machine translation and speech synthesis for Judeo-Spanish. First, we developed a rule-based machine translator from Spanish to Judeo-Spanish. This base translator was used to apply a data augmentation technique. Second, we developed three bidirectional machine translation models between Judeo-Spanish and Spanish, Turkish and English, being the first neural-based systems for this language. Although some of our models do not perform optimally, we believe that this work is the basis for future research regarding this language, as well as motivating research for extremely low-resourced languages using data augmentation strategies. Third, we developed speech synthesis models for Judeo-Spanish, achieving an acceptable result by fine-tuning on an English model. Data, model checkpoints, development and test sets and configuration files are shared openly on project's data portal Ladino Data Hub and our Github repository. Finally, we created a web-application for machine translation with voice to help language learners, researchers and linguists who want to study Judeo-Spanish.

\section{Acknowledgements}

This paper was written as part of the project ``Judeo-Spanish: Connecting the two ends of the Mediterranean'' carried out by Col$\cdot$lectivaT and Sephardic Center of Istanbul within the framework of the ``Grant Scheme for Common Cultural Heritage: Preservation and Dialogue between Turkey and the EU (CCH-II)'' implemented by the Ministry of Culture and Tourism of the Republic of Türkiye with the financial support of the European Union. The content of this paper is the sole responsibility of the authors and does not necessarily reflect the views of the European Union.

Yasmin Moslem contributed to this work while she was pursuing her PhD degree, supported by the \mbox{Science} Foundation Ireland Centre for Research Training in Digitally-Enhanced Reality (d-real) under Grant No. 18/CRT/6224, and the ADAPT Centre for Digital Content Technology which is funded under the Science Foundation Ireland (SFI) Research Centres Programme (Grant No. 13/RC/2106) and is co-funded under the European Regional Development Fund.

We would like to thank the excellent volunteer work by Brian Russell, who helped us digitize \textit{Fraza del dia} content.

\section{Bibliographical References}\label{reference}

\bibliographystyle{lrec2022-bib}
\bibliography{lrec2022-example}

\end{document}